\documentclass[10pt,twocolumn,letterpaper]{article}
\usepackage[pagenumbers]{cvpr}              
\usepackage[accsupp]{axessibility} 

\makeatletter
\@namedef{ver@everyshi.sty}{}
\makeatother

\usepackage{graphicx}
\usepackage{amsmath}
\usepackage{amssymb}
\usepackage{booktabs}
\usepackage{hyperref}
\hypersetup{colorlinks}
\usepackage{todonotes}
\usepackage{xcolor}
\usepackage{caption}
\usepackage{subcaption}
\usepackage{mathtools}
\usepackage{svg}
\usepackage{stackengine}
\usepackage{siunitx}
\definecolor{reco_orange}{RGB}{255, 191, 128}
\definecolor{restorer_blue}{RGB}{204, 204, 255}
\definecolor{aruco_green}{RGB}{128, 255, 128}
\definecolor{segmentation_gray}{RGB}{204, 204, 204}
\newboolean{long_appendix}
\setboolean{long_appendix}{false}

\usepackage{tikz}
\usepackage{pgfplots}
\usepackage{array,multirow,graphicx}
\usetikzlibrary{shapes,arrows, automata}
\usepackage{float}
\usetikzlibrary{scopes}
\usetikzlibrary{backgrounds}
\usetikzlibrary {shadows} 
\pgfdeclarelayer{edgelayer}
\pgfdeclarelayer{nodelayer}
\pgfsetlayers{edgelayer,nodelayer,main}

\tikzstyle{none}=[shape=circle, fill=none, draw=none]
\tikzstyle{Input}=[trapezium, drop shadow={opacity=0.25}, draw, fill={orange!50}, trapezium left angle=70, trapezium right angle=110, trapezium stretches body, text width={\pgfkeysvalueof{/pgf/minimum width}-2*\pgfkeysvalueof{/pgf/inner xsep}}, text centered, minimum width=5em, minimum height=3em, node distance=3cm]
\tikzstyle{Processor}=[fill={orange!50}, drop shadow={opacity=0.25}, draw=black, rounded corners, shape=rectangle, text centered, text width=6em, minimum height=4em]
\tikzstyle{Processor2}=[fill={blue!20}, drop shadow={opacity=0.25}, draw=black, rounded corners, shape=rectangle, text width=6em, text centered, minimum height=4em]
\tikzstyle{Input2}=[trapezium, drop shadow={opacity=0.25}, draw, fill={blue!20}, trapezium left angle=70, trapezium right angle=110, trapezium stretches body, text width=7em, text centered, minimum width=5em, minimum height=3em, node distance=3cm]
\tikzstyle{Input3}=[trapezium, drop shadow={opacity=0.25}, draw, fill={green!50}, trapezium left angle=70, trapezium right angle=110, trapezium stretches body, text width=7em, text centered, minimum width=5em, minimum height=3em, node distance=3cm]
\tikzstyle{Process3}=[fill={green!50}, drop shadow={opacity=0.25}, draw=black, rounded corners, shape=rectangle, text width=8em, text centered, minimum height=4em]
\tikzstyle{Input4}=[trapezium, drop shadow={opacity=0.25}, draw, fill={black!20}, trapezium left angle=70, trapezium right angle=110, trapezium stretches body, text width=7em, text centered, minimum width=5em, minimum height=3em, node distance=3cm]
\tikzstyle{Process4}=[fill={black!20}, drop shadow={opacity=0.25}, draw=black, rounded corners, shape=rectangle, text width=8em, text centered, minimum height=4em]
\tikzstyle{Process5}=[fill=white, drop shadow={opacity=0.25}, draw=black, rounded corners, shape=rectangle, text width=12em, text centered, minimum height=2em]
\tikzstyle{Overview1}=[fill={orange!50}, drop shadow={opacity=0.25},draw=black, shape=rectangle, text width=12em, text centered]
\tikzstyle{Overview2}=[fill={blue!20}, drop shadow={opacity=0.25},draw=black, shape=rectangle, text width=12em, text centered]
\tikzstyle{Overview3}=[fill={green!50}, drop shadow={opacity=0.25},draw=black, shape=rectangle, text width=12em, text centered]
\tikzstyle{Overview4}=[fill={black!20}, drop shadow={opacity=0.25},draw=black, shape=rectangle, text width=12em, text centered]

\tikzstyle{arrow_head}=[->, drop shadow={opacity=0.25}, line width=0.5mm]
\tikzstyle{default}=[-, drop shadow={opacity=0.25}, line width=0.5mm]

\usepackage[capitalize]{cleveref}
\crefname{section}{Sec.}{Secs.}
\Crefname{section}{Section}{Sections}
\Crefname{table}{Table}{Tables}
\crefname{table}{Tab.}{Tabs.}

\newcommand\norm[1]{\left\lVert#1\right\rVert}

\usepackage{scalerel}
\usetikzlibrary{svg.path}
\definecolor{orcidlogocol}{HTML}{A6CE39}
\tikzset{
  orcidlogo/.pic={
    \fill[orcidlogocol] svg{M256,128c0,70.7-57.3,128-128,128C57.3,256,0,198.7,0,128C0,57.3,57.3,0,128,0C198.7,0,256,57.3,256,128z};
    \fill[white] svg{M86.3,186.2H70.9V79.1h15.4v48.4V186.2z}
                 svg{M108.9,79.1h41.6c39.6,0,57,28.3,57,53.6c0,27.5-21.5,53.6-56.8,53.6h-41.8V79.1z M124.3,172.4h24.5c34.9,0,42.9-26.5,42.9-39.7c0-21.5-13.7-39.7-43.7-39.7h-23.7V172.4z}
                 svg{M88.7,56.8c0,5.5-4.5,10.1-10.1,10.1c-5.6,0-10.1-4.6-10.1-10.1c0-5.6,4.5-10.1,10.1-10.1C84.2,46.7,88.7,51.3,88.7,56.8z};
  }
}
\newcommand\orcid[1]{\href{https://orcid.org/#1}{\mbox{\scalerel*{
\begin{tikzpicture}[yscale=-1,transform shape]
\pic{orcidlogo};
\end{tikzpicture}
}{|}}} \href{https://orcid.org/#1}{}}

\def\confName{CVPR}
\def\confYear{2023}

\begin{document}

\title{CherryPicker: Semantic Skeletonization and Topological Reconstruction of Cherry Trees}
\author{Lukas Meyer\textsuperscript{1}\orcid{0000-0003-3849-7094}, Andreas Gilson\textsuperscript{2}\orcid{0009-0001-1674-0447}, Oliver Scholz\textsuperscript{2}\orcid{0000-0002-6304-2182}, Marc Stamminger\textsuperscript{1}\orcid{0000-0001-8699-3442}\\
\textsuperscript{1}Visual Computing, Friedrich-Alexander-Universität Erlangen-Nürnberg, Germany\\
\textsuperscript{2}Fraunhofer Institute for Integrated Circuits (IIS), Erlangen, Germany\\
{\tt\small lukas.meyer@fau.de}
}

\maketitle
\begin{abstract}
In plant phenotyping, accurate trait extraction from 3D point clouds of trees is still an open problem. For automatic modeling and trait extraction of tree organs such as blossoms and fruits, the semantically segmented point cloud of a tree and the tree skeleton are necessary. 
 Therefore, we present CherryPicker, an automatic pipeline that reconstructs photo-metric point clouds of trees, performs semantic segmentation and extracts their topological structure in form of a skeleton. Our system combines several state-of-the-art algorithms to enable automatic processing for further usage in 3D-plant phenotyping applications. 
Within this pipeline, we present a method to automatically estimate the scale factor of a monocular reconstruction to overcome scale ambiguity and obtain metrically correct point clouds. 
Furthermore, we propose a semantic skeletonization algorithm build up on Laplacian-based contraction. We also show by weighting different tree organs semantically, our approach can effectively remove artifacts induced by occlusion and structural size variations. 
CherryPicker obtains high-quality topology reconstructions of cherry trees with precise details. 

\end{abstract}

\section{Introduction}

Digital plant phenotyping is a vital tool in the crop improvement process for breeders and farmers as it yields objective and precise plant traits to use for a variety of purposes. For instance, it may help to find the breed with the highest resistance to drought stress or help a farmer determine the optimum time for the harvest to maximize yields. It is thus an increasingly important tool to help to enlarge crop production for a growing world population despite the worsening conditions caused by climate change. Plant phenotyping employs the use of sensors, cameras, and other digital tools to collect data at plant level, field level, or even from space. While field-scale phenotyping can generate information about large crop areas in a short time, plant-level phenotyping is the only way to extract detailed information about each individual plant and every single organ of interest. This data - in connection with additional environmental information e.g. soil conditions - supports informed decision-making regarding breeding, irrigation, pest control, or harvest. Moreover, controlled agricultural systems enable the automation of tasks such as pruning \cite{automated_pruning} and harvesting \cite{automated_harvesting}. 
\begin{figure}[t]
    \centering
    \includegraphics[width=0.98\linewidth, trim=0 15 0 5,clip]{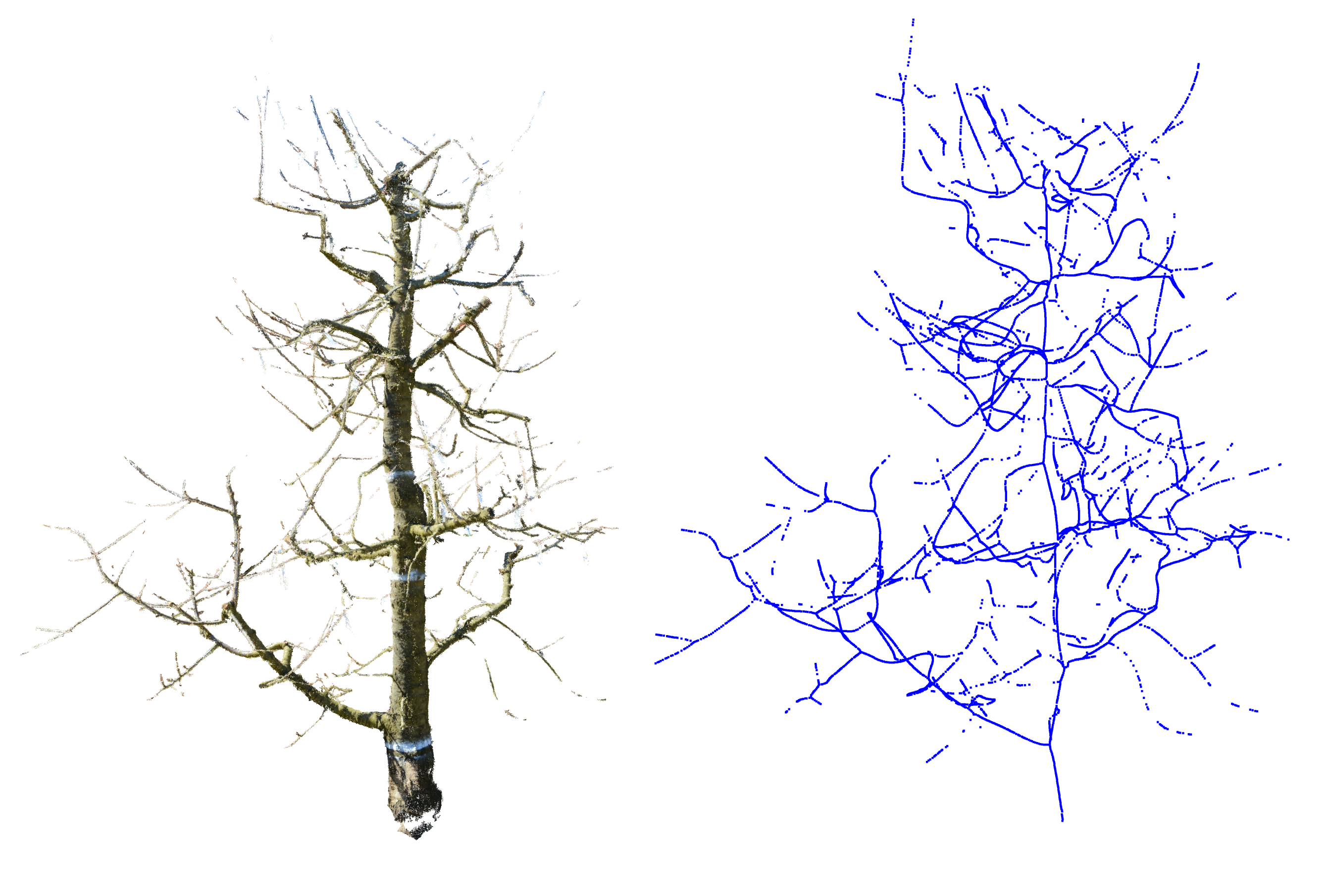}
    \caption{Visualization of the reconstructed point cloud $\mathbf{C}_b$ on the left. It has been processed by \textit{CherryPicker} from a set of images $\mathcal{I}$, aligned in space, de-noised, and segmented into both classes trunk and branch. On the right is the extracted skeleton from our proposed semantic Laplacian-based contraction algorithm.  }
    \label{fig:aruco_scale_estimation}
\end{figure}
Significant advances were made recently, particularly in the field of plant phenotyping and precision farming. The use of terrestrial laser scanning \cite{Tsoulias2D, lidar_backpack} or photo-metric reconstruction techniques \cite{PhotogrammetryReconstruction, tree_dynamic, Straub22} for tree modeling is an emerging research field. The focus of these studies is to accurately quantify the intrinsic parameters of trees, particularly geometry, photo-metric shape, and topological structure.

A tree's geometric and photo-metric representation can either be used for visual representation or transformed into a topological structure using skeletonization algorithms \cite{campino, rosa, Laplacian, L1}. 
In this way, annual shoots, branches, and intersections can be identified unambiguously. This enables estimation of inter- and intra-seasonal growth of trees regarding the increase in length and thickness. 
In addition, spatial information of the location of annual shoots, branches, buds, leaves, flowers, and fruits within a tree can be located and individually tracked during their vegetative growth. 
With \textit{CherryPicker}, we present a pipeline that extracts the structure of cherry trees in form of skeletons, as depicted in Fig. \ref{fig:aruco_scale_estimation}. This enables straightforward navigation within a complex network of branches. 
While the presentation of our work is limited to the domain of cherry trees, the proposed methods can be applied to trees in general or other specific domains with similar-looking data, e.g apple trees or cotton \cite{saeed2022optimal}. 
In summary, we make the following contributions:
\begin{itemize}
    \item We present an automatic reconstruction pipeline for extracting the topology of cherry trees from photo-metric data. 
    \item We extend the Laplacian-based contraction algorithm \cite{Laplacian} by using semantic point cloud information of cherry trees and the code is made open-source\footnote{PC-Skeletor: \href{https://github.com/meyerls/pc-skeletor}{github.com/meyerls/pc-skeletor}}.
    \item We present an ArUco-marker-based scale factor estimation to automatically overcome the missing scale factor for dense and sparse point clouds. The code is made open-source\footnote{Aruco-Estimator: \href{https://github.com/meyerls/aruco-estimator}{github.com/meyerls/aruco-estimator}}. 
\end{itemize}

\section{Related Work}
\noindent \textbf{Tree Reconstruction}. Photo-metric tree reconstructions have become increasingly popular in recent years. \cite{PhotogrammetryReconstruction} generated a 3D point cloud from a severed and defoliated tree to obtain digital ground truth. In a laboratory setup \cite{tree_dynamic} created a 3d reconstruction of a fruit tree in order to be able to model its dynamic behavior. Using a deep learning approach \cite{DeepTreeReconstruction} reconstruct the tree structure from a single image using a conditional generative adversarial network. Similar to our approach, \cite{Straub22} presents a pipeline structure that creates a graph representation from photo-metric 3D point clouds of meadow orchard trees. Drawbacks of this pipeline are the manual point cloud scaling, manual sky silhouette removal, and the removal of detailed tree structures.\\ 

\noindent\textbf{Scale Factor Estimation}. There are different approaches for determining the scale factor of a monocular 3D reconstruction. \cite{ScaleFactorEstimation} covers the method of placing a calibration object into the scene and manually scaling the model or by taking different poses with known relative displacement.
Automated estimation of the scale factor is not well studied in the literature, and further limited in its availability as an open-source package. Only the reconstruction software AliceVision \cite{AliceVision} mentions using CCtags\cite{cctags} as a method to obtain metric reconstructions.\\

\noindent\textbf{Skeletonization}. Many methods for extracting skeletons from implicit (e.g. signed distance fields) and explicit (point clouds and polygon meshes) representations are well-studied in the literature. A comprehensive overview of state-of-the-art algorithms to extract 3D skeletons is provided in \cite{skeletonSOTA}. 

The CAMPINO algorithm \cite{campino} processes a point cloud by partitioning the set of points using an octree structure. Drawbacks of this approach are the required manual selection of a proper octree cell size to avoid topological faults and the susceptibility to incomplete point clouds. The latter was improved in the follow-up work SkelTre\cite{skeltre} that extracts topologically correct skeletons from incomplete point clouds by taking shape approximations into account.
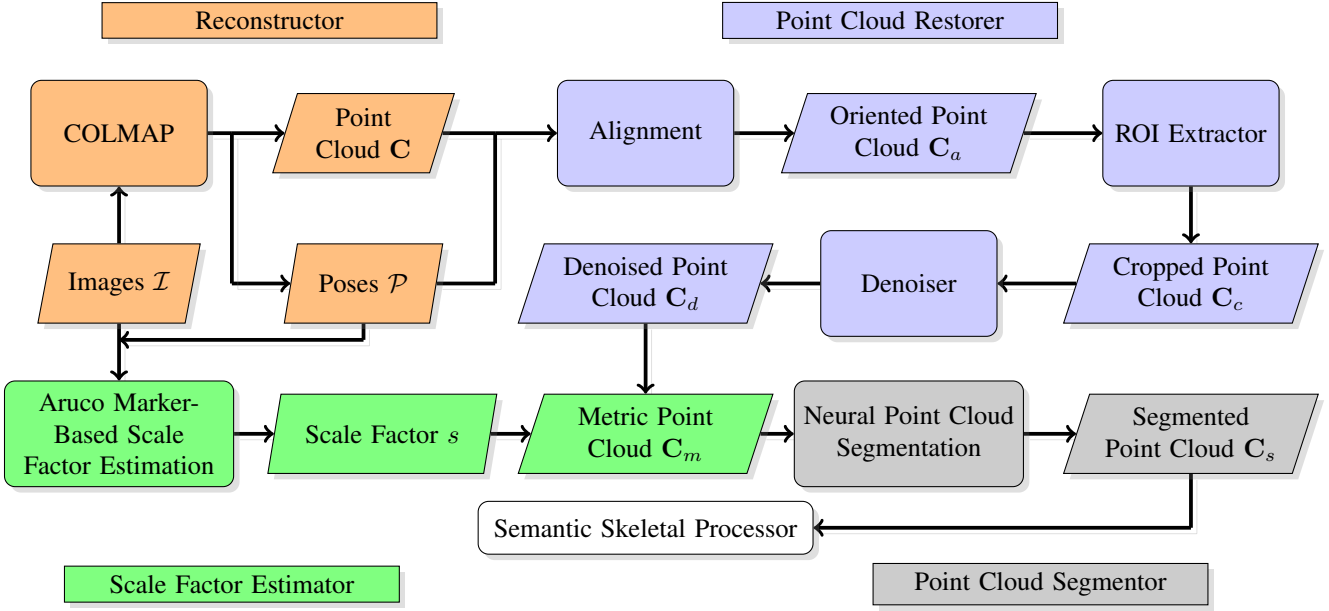
\begin{figure*}[ht]
  \begin{tikzpicture}
	\begin{pgfonlayer}{nodelayer}
		\node [style=Input] (1) at (-12, 0) {Images $\mathcal{I}$};
		\node [style=Processor] (2) at (-12, 2) {COLMAP};
		\node [style=Input] (3) at (-8.75, 2) {Point Cloud $\mathbf{C}$};
		\node [style=Input] (4) at (-8.75, 0) {Poses $\mathcal{P}$};
		\node [style=Processor2] (7) at (-5, 2) {Alignment};
		\node [style=Processor2] (8) at (2.25, 2) {ROI Extractor};
		\node [style=Processor2] (9) at (-1.5, 0) {Denoiser};
		\node [style=Input2] (11) at (2.25, 0) {Cropped Point Cloud $\mathbf{C}_c$};
		\node [style=Input2] (12) at (-5, 0) {Denoised Point Cloud $\mathbf{C}_d$};
		\node [style=none] (13) at (-10.5, 2) {};
		\node [style=none] (14) at (-10.5, 0) {};
		\node [style=Input2] (16) at (-1.5, 2) {Oriented Point Cloud $\mathbf{C}_a$};
		\node [style=none] (17) at (-7, 0) {};
		\node [style=none] (18) at (-7, 2) {};
		\node [style=Process3] (19) at (-12, -2) {Aruco Marker-Based Scale Factor Estimation};
		\node [style=Input3] (20) at (-8.5, -2) {Scale Factor $s$};
		\node [style=none] (22) at (-12, -0.75) {};
		\node [style=Input3] (23) at (-5, -2) {Metric Point Cloud $\mathbf{C}_m$};
		\node [style=none] (24) at (-8.75, -0.75) {};
		\node [style=Process4] (25) at (-1.5, -2) {Neural Point Cloud Segmentation};
		\node [style=Input4] (26) at (2.25, -2) {Segmented Point Cloud  $\mathbf{C}_s$};
		\node [style=none] (28) at (2.25, -3.25) {};
		\node [style=Overview1] (29) at (-10, 3.5) {Reconstructor};
		\node [style=Overview2] (30) at (-1.75, 3.5) {Point Cloud Restorer};
		\node [style=Overview3] (31) at (-10.5, -4) {Scale Factor Estimator};
		\node [style=Overview4] (32) at (0.25, -4) {Point Cloud Segmentor};
		\node [style=Process5] (34) at (-5, -3.25) {Semantic Skeletal Processor};
	\end{pgfonlayer}
	\begin{pgfonlayer}{edgelayer}
		\draw [style={arrow_head}] (1) to (2);
		\draw [style={arrow_head}] (13.center) to (3);
		\draw [style={arrow_head}] (14.center) to (4);
		\draw [style={arrow_head}] (7) to (16);
		\draw [style={arrow_head}] (16) to (8);
		\draw [style={arrow_head}] (8) to (11);
		\draw [style={arrow_head}] (11) to (9);
		\draw [style={arrow_head}] (9) to (12);
		\draw [style=default] (2) to (13.center);
		\draw [style=default] (13.center) to (14.center);
		\draw [style=default] (4) to (17.center);
		\draw [style={arrow_head}] (18.center) to (7);
		\draw [style=default] (3) to (18.center);
		\draw [style=default] (17.center) to (18.center);
		\draw [style={arrow_head}] (1) to (19);
		\draw [style={arrow_head}] (19) to (20);
		\draw [style={arrow_head}] (20) to (23);
		\draw [style={arrow_head}] (12) to (23);
		\draw [style=default] (4) to (24.center);
		\draw [style={arrow_head}] (24.center) to (22.center);
		\draw [style={arrow_head}] (23) to (25);
		\draw [style={arrow_head}] (25) to (26);
		\draw [style=default] (26) to (28.center);
		\draw [style={arrow_head}] (28.center) to (34);
	\end{pgfonlayer}
\end{tikzpicture}
  \caption{System Pipeline of \textit{CherryPicker}. The pipeline is separated into six different working blocks \textcolor{reco_orange}{Reconstructor}, \textcolor{restorer_blue}{Point Cloud Restorer}, \textcolor{aruco_green}{Scale Factor Estimator}, \textcolor{segmentation_gray}{Point Cloud Segmentor} and Semantic Skeletal Processor. From a set of images $\mathcal{I}$ a dense point cloud is reconstructed and afterward aligned, denoised, scaled, and segmented into different tree organs. With the information obtained from the segmentation, it is possible to add semantic weighting to the Laplacian-based skeletonization algorithm.}
  \label{fig:system_pipeline}%
\end{figure*}

ROSA \cite{rosa} extracts curved skeletons from an oriented point cloud in cases where the underlying shape is roughly cylindrical. A strength of this algorithm is that, given the point normals, it is able to make assumptions about the cylindrical shape of point clouds with missing data. 
The L1-Medial Skeleton algorithm \cite{L1} is another promising approach for skeletonization on point clouds with significant noise, outliers, and large areas of missing data. The Point2Skeleton Network \cite{Point2Skeleton} is an unsupervised deep learning approach to extract a generalized skeletal representation.
Laplacian-based contraction \cite{Laplacian} is a method to extract a curved skeleton of a point cloud. It iteratively applies Laplacian smoothing and also maintains the global shape characteristics. 
Our skeletonization algorithm is a modification of Laplacian-based contraction and differs in its semantic components by weighting different classes of a point cloud according to their structural properties.

\section{\textit{CherryPicker} System Pipeline}

This chapter covers the proposed system pipeline as depicted in Fig. \ref{fig:system_pipeline}.

\subsection{Reconstructor}
\label{ssec:data_acquisition}
The recording of the cherry trees was conducted at a research orchard facility, in \textit{Franconian Switzerland}, Germany, which has a wide range of cherry trees in different age classes. For our measurement campaign, multiple cherry trees were scanned and the pipeline will be demonstrated using a single tree's data set. 

Image acquisition was performed using a Nikon D7100 DSLR camera \cite{NikonManual} equipped with a lens with a focal length of 35\,mm. Time of day of the acquisition was between sunrise and noon. In general, the ambient conditions for all images can be described as slightly cloudy to sunny, almost windless, and thus without undesirable object motion. Images were collected from both sides of the trees at a constant distance at different heights. On average, around 250 images were acquired for every single tree.
The input of our system pipeline is a set of images $\mathcal{I} = \{I_i \, | \, i = 1,...,N_I \}$, where $N_I $ indicates the total number of images. We use state-of-the-art open-source reconstruction software COLMAP\cite{schoenberger2016sfm, schoenberger2016mvs}. It extracts the camera poses $\mathcal{P} = \{\mathbf{P}_i \in SE(3) \, | \, i = 1,...,N_I \}$ and a dense point cloud  $\mathbf{C} \in \mathbb{R}^{N_C \times 3}$, where $N_C $ indicates the number of points. Both,  camera poses and the dense point clouds, were used for subsequent processing.

\subsection{Point Cloud Restorer}
\label{ssec:pc_restorer}
The resulting point cloud $\mathbf{C}$ is not oriented in space, is arbitrarily scaled, and contains points of background structures as well as noise. Therefore, the restoration step, depicted as blue blocks in Fig. \ref{fig:system_pipeline}, aims to transform the point cloud to a defined orientation, extract the region of interest, and perform noise reduction. 
\subsubsection{Alignment}
The objective of the alignment process is to rearrange the point cloud $\mathbf{C}$ and poses $\mathcal{P}$ in order to achieve a defined orientation. 
In other words, we aim to align the ground plane of the reconstruction with the $xy$-plane in the local coordinate system.
This can be achieved by computing the rotation matrix between the ground normal vector to the normal vector of the $xy$-plane. To determine the normal vector of the ground plane, we utilized the RANSAC algorithm\cite{leonardo_mariga_2022_7212568} to fit a plane in Cartesian form into the given point cloud (during the outdoor reconstruction the background gets densely represented). Afterward we determine the rotation by utilizing Rodrigues rotation formula \cite{szeliski2022computer} and translation between the ground plane's normal vector and the normal vector of the $xy$-plane. 
To align the local tree coordinate system with our global coordinate system it is necessary to compute the $z$-direction of the tree relative to the $xy$-plane. 
Therefore, we determine the center of mass of the point cloud, and if the center of mass lies in the negative $z$-direction a $180^\circ$ rotation is applied and we obtain an aligned point cloud $\mathbf{C}_a$.

\subsubsection{Region of Interest}

The purpose of extracting the region of interest (ROI) is twofold: to remove the background that appears in the outdoor reconstruction and to isolate the tree itself.
To extract the ROI from the oriented point cloud $\mathbf{C}_a$ it is assumed that all camera positions are positioned around the tree and that a bounding box around the camera origins encompasses the entire tree.  To achieve this, the positions of the camera origin $\mathcal{O} = \{ t^{\text{Cam}}_i | i = 1, ..., N_I \}$ are utilized to create an axis-aligned bounding box for the $xy$-direction that contains only the points of the actual tree, resulting in a cropped point cloud denoted as $\mathbf{C}_c$. 

\subsubsection{Denoising}
 
The extracted tree point cloud of $\mathbf{C}_c$ still contains erroneous points originating from different sources. 
The first source of error are noisy depth values caused by an erroneous depth estimation, which are manifested in the reconstructed point cloud as isolated points scattered randomly in space. 
To mitigate this issue, a statistical filter is employed to detect and eliminate these points based on the number of neighboring points within a range defined by the standard deviation of the average Euclidean distances between all pairs of points \cite{open3d}.
The second type of error is caused by images taken against the sky.
Due to coarse depth maps based on the selected window size of the stereo matching algorithm, sky image colors are projected into the reconstructed branches of the point clouds. 
Especially thin structures such as annual shoots are particularly vulnerable to this error. 
This phenomenon will be referred to as a sky silhouette in the remainder of this paper.

The appearance of the sky in images can vary depending on factors such as the direction of the sun, changes in lighting throughout the day, and changing weather conditions. 
Based on \cite{TheUseofEuclideanGeometricDistanceonRGBColorSpacefortheClassificationofSkyandCloudPatterns}, it can be stated that the patterns of the sky and clouds are usually located within a specific subset of the RGB color space.
To remove the sky's silhouette from the reconstructed 3D point clouds a U-Net\cite{unet} based neural network was implemented. 
The network separates the sky from the ground and afterward the RGB sky pixels are clustered on their appearance, which subsequently allows the identification of dominant sky colors. 

For U-Net we chose a similar approach as \cite{SkySeg} by using the Scene UNderstanding (SUN) dataset \cite{SUNDataset} and searching for images with class labels \textit{sky}. 
This way $\sim5000$ images were gathered and then resized to $400 \text{px} \times 400 \text{px}$. Using multiple augmentation methods, in particular scaling, translation, and elastic transform, the dataset was artificially increased and split into $80\%$ training data and $20 \%$ validation data. The network was trained with a batch size of $3$ over $15$ epochs, to minimize binary cross-entropy loss using Adam \cite{Adam} as an optimizer and a decaying learning rate starting at $10^{-5}$. 
The resulting model achieved a DICE loss of $0.830$ on the validation dataset. 
The classes sky and ground were predicted both with an F1-score of $0.877$. 

Using this network, we are able to extract the color values of the detected sky from the image set $\mathcal{I}$. According to \cite{TheUseofEuclideanGeometricDistanceonRGBColorSpacefortheClassificationofSkyandCloudPatterns}, the color values of the sky and clouds should exhibit a linear behavior and gather around the RGB diagonal. We found that in several image sets, the color pattern curved along the blue wall of the RGB cube. 
To extract the dominant color range of the sky and to remove points with similar color values, we utilized Density-Based Spatial Clustering (DBSCAN)\cite{DBSCAN}. Since clustering does not work on relatively large discretization distances between color values, we added $\mathcal{N}(0, \frac{1}{256})$ to the color values beforehand.

\subsection{Scale Factor Estimator}
\label{ssec:scalefactorestimation}

The determination of the scale factor is an essential part of our application. 
Therefore, we elaborate in this section on how to specify the scale factor to obtain a metric 3D reconstruction automatically. 
In the literature only \cite{AliceVision} mentions using CCTags for scaling and orientation of the scene automatically. 
Therefore we developed an automatic scale factor estimation, where only a single ArUco marker with a known size has to be placed within the scene to determine the scale factor.

Monocular multi-view stereo (MVS) suffers from scale ambiguity because monocular systems are unable to determine the true scale of a scene. This is rooted in the underlying epipolar geometry, which characterizes the mathematical relationship between two images of a scene captured from different viewpoints. This relationship can be described using the Essential Matrix. It is a $3\times3$ matrix that connects two image planes in a stereo pair, encoded with five degrees of freedom (DoF). These include relative rotation (3 DoF), translation (2 DoF), and intrinsic camera parameters \cite{szeliski2022computer}. However, translation has only 2 degrees of freedom and thus describes only the direction of translation between the two cameras and not the scale.

The tree scene reconstructed from SfM (Structure from Motion) is described by a set of input images $\mathcal{I}$ and the computed poses $\mathcal{P}$. The set $\mathcal{I}$ is scanned for markers and outputs a set of images $\mathcal{M} = \{I_j \, | \, j = 1,...,N_J \}$ with detected markers whereby $N_J$ indicates the number of images containing an ArUco marker. However, as not every image contains an ArUco marker the minimum requirement for our scale factor estimation is $N_J \geq 2$. 

The detected 2D location of an ArUco marker within an image $I_j$ is a set of four corner points $\mathbf{a}_j = \{ \mathbf{c_1} , \mathbf{c_2} , \mathbf{c_3} , \mathbf{c_4} \}$ with $\mathbf{c}_k \in \mathbb{R}^{2}$ with $k = 1,2,3,4$. Given the camera origin $\mathbf{t}_j$ and the unit direction vector $\mathbf{u}_{jk}$, pointing to the corner of an ArUco marker, a line can be cast through each of the 2D corners $\mathbf{c}_k$ for every image $I_j$ by
\begin{equation}
    \mathbf{r}_{jk}(\lambda) = \mathbf{t}_j +  \lambda \mathbf{u}_{jk}.
\end{equation}
The direction vector of the line is defined as the normalized vector
\begin{equation}
     \mathbf{u}_{jk} = \mathbf{R}_j \frac{\mathbf{K}^{-1}\mathbf{\widetilde{c}}_{jk}}{\norm{\mathbf{K}^{-1}\mathbf{\widetilde{c}}_{jk}}_2}.
\end{equation}
$\mathbf{K} \in \mathbb{R}^{3 \times 3}$ is denoted as the intrinsic camera matrix that is identical for all images. $\widetilde{\mathbf{c}}_{jk} = (\mathbf{c}_{jk} \quad 1 )^\top$ is the 2D image coordinate represented using homogeneous coordinates. $\mathbf{R}_j \in SO(3)$ is the rotational part of the camera extrinsic $\mathbf{P}_j$.

Finally, we can group from the 3D lines $\mathcal{L}_k= \{\mathbf{r}_{jk} \, | \, j = 1...N_J \}$ from all images in the set $\mathcal{M}$ that intersect in the same corner of the ArUco marker. Ideally the intersection point $\mathbf{x} \in \mathbb{R}^{3}$ can be computed by setting up a  system of linear equations and solving it. Since all measurements are noisy, it can be assumed that the lines do not intersect at a unique point. Instead, we acquire the optimal solution in the least-square sense by minimizing the sum of squared distances\cite{intersectionoflines}. The objective of this method involves finding a feasible point $\mathbf{x}$ that minimizes the sum of the squared distances from the point $\mathbf{x}$ to each line.
The squared orthogonal distance from an arbitrary point $\mathbf{p}$ to a line is given by
\begin{equation}
    \begin{aligned}
        D(\mathbf{x}; \mathbf{t}, \mathbf{u}) =  \norm{(\mathbf{t} - \mathbf{x}) - ((\mathbf{t} - \mathbf{x})^\top \mathbf{u})\mathbf{u}}^2_2.
    \end{aligned}
\end{equation}
By using all lines from our subset $\mathcal{L}_k$ for one corner $k$ the unique solution can be found by minimizing the sum of squared distances of all corresponding lines with
\begin{equation}
    \begin{aligned}
        D(\mathbf{x}_k; \mathbf{T}, \mathbf{U}_k) &=  \sum^{|\mathcal{L}_k|}_j D(\mathbf{x}_k; \mathbf{t}_j, \mathbf{u}_{jk}),
    \end{aligned}
    \label{eq:objective}
\end{equation}
where $\mathbf{T} \in \mathbb{R}^{|\mathcal{L}_k| \times 3}$ describes the camera origin for all cameras observing an ArUco marker and $\mathbf{U}_k \in \mathbb{R}^{|\mathcal{L}_k| \times 3}$ is the direction vector to the identical ArUco marker corner across all involved images. With Eq. \ref{eq:objective} as an objective function we can minimize this function
to find a feasible point $\mathbf{x}_k \in \mathbb{R}^{3}$ for one corner point.
By solving the minimization problem in the least squares sense for all four corner points we obtain a set $\{ \mathbf{x}_1, \mathbf{x}_2, \mathbf{x}_3, \mathbf{x}_4 \}$ of the ArUco marker corners in 3D space. Set $\mathcal{A} = \{ (\mathbf{x}_1, \mathbf{x}_2), (\mathbf{x}_2, \mathbf{x}_3), (\mathbf{x}_3, \mathbf{x}_4), (\mathbf{x}_4, \mathbf{x}_1) \}$ describes all pairs of neighboring corners. Thus, the mean distance between neighboring corners can be calculated by
\begin{equation}
    \bar{d}_{\text{scene}} = \frac{1}{4}\sum_{(\mathbf{x}_i, \mathbf{x}_j) \in \mathcal{A}} \norm{\mathbf{x}_i - \mathbf{x}_j}_2 .
\end{equation}
Since the distance of the square ArUco marker $d_{\text{\text{aruco}}}$ is known the scale factor $s$ can be computed by $s = d_{\text{aruco}}/\bar{d}_{\text{scene}}$.
Finally, the set of poses $\mathcal{P}$ and the denoised point cloud $\mathbf{C}_d$ are scaled by $s$ such that we obtain a set of metric poses $\mathcal{P}_m$ and the metric point cloud $\mathbf{C}_m$. In a preceding analysis \cite{GIL_For5G} we showed that estimating the diameter of the trunk (indicated by a white ring) at three different locations is obtained with an accuracy of up to $6$\,mm.
\subsection{Point Cloud Segmentor}
\label{ssec:pcsegmentation}
\begin{figure}[h]
    \centering
    \includegraphics[width=0.7\linewidth]{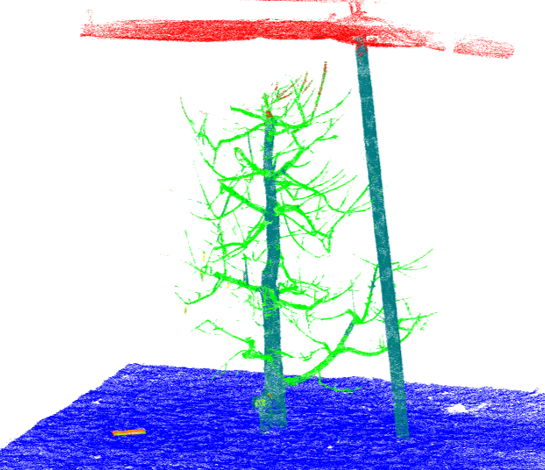}%
    \caption{Visualization of the segmented point cloud $\mathbf{C}_s$ from the validation dataset. The prediction classes are ground, trunk, branches, signs, markers, calibration units, and roof. It is noted that the supporting bar is annotated in the trunk class and separated afterward.}%
    \label{fig:pc_segmented}%
\end{figure}

In this section, we briefly elaborate on the segmentation of the metric point cloud $\mathbf{C}_m$. The goal is to segment the cherry trees into relevant semantic categories to be able to differentiate between their plant organs. 
The aligned, denoised, and scaled point clouds of 16 annotated tree scans were used to create a dataset for the application of 3D semantic segmentation via deep learning. 
To account for the varying sizes and densities the point clouds are down-sampled to 4 million points each and then sub-sampled using a slightly modified version of spherical sub-sampling \cite{sphericalsubsampling}. 
Every resulting data point contains RGB values as well as global and normalized $xyz$-coordinates that we used to train a Deep Graph Convolutional Neural Network (DGCNN) \cite{DGCNN} in order to map each individual point to a corresponding target class. 
As network architecture, we used the Pytorch-geometric \cite{PyTorchGeometric} implementation of DGCNN, which interprets point clouds as graph networks to recover their topological information that is further processed by a specifically designed operator called EdgeConv \cite{DGCNN}. Our best model was reached after training 35 epochs with a batch size of 16 sub-spheres per step, with Adam as an optimizer, negative log-likelihood loss, and a decaying learning rate starting at 0.003.
The resulting network was able to predict 7 different target classes (ground, trunk, branches, signs, marker, calibration units, and roof) with a total IoU of $0.95$ on validation data. The most relevant categories for our purpose: branch and trunk were predicted both with an F1-score of $0.93$. The result is a segmented point cloud $\mathbf{C}_s$ as displayed in Fig. \ref{fig:pc_segmented}.
\section{Skeleton and Topology}
In this chapter, we aim to extract the skeleton of the processed point cloud and transform it into a simplified graph.
\subsection{Point Cloud Contraction}
\label{ssec:pc_contraction}
We employed Laplacian-based contraction (LBC)\cite{Laplacian} as the skeletonization algorithm. This algorithm is a geometric skeletonization method that compresses a 3D input point cloud $\mathbf{C}\in \mathbb{R}^{N_C \times 3}$, with $N_C$ being the number of points of the tree, to a zero-volume point set. The contraction process preserves the input model's global geometric characteristics by anchoring points and repeatedly applying Laplacian smoothing. By solving the linear system of equations  
\begin{equation}
\begin{bmatrix}
\mathbf{W}_L \mathbf{L}\\
\mathbf{W}_H
\end{bmatrix} \mathbf{C}^{'} =
\begin{bmatrix}
\mathbf{0}\\
\mathbf{W}_H \mathbf{C}
\end{bmatrix}    
\label{eq:lbc}
\end{equation}
iteratively the contracted point cloud $\mathbf{C}^{'}$ can be computed. $\mathbf{W}_L \in \mathbb{R}^{N_C \times N_C}$ and $\mathbf{W}_H \in \mathbb{R}^{N_C \times N_C}$ are sparse diagonal weighting matrices to regulate the contraction and attraction constraints respectively. $\mathbf{L} \in \mathbb{R}^{N_C \times N_C}$ is defined as cotangent Laplacian matrix. By repeatedly solving the linear system of equations, the weight matrices $\mathbf{W_L}$ and $\mathbf{W_H}$ are updated, and $\mathbf{L}$ is recalculated. This process is repeated until the termination criterion is reached. 
\subsection{Semantic Point Cloud Contraction}
\label{ssec:semantic_pc_contraction}
Standard LBC is prone to mal-contraction in cases where there is a significant disparity in diameter between trunk and branches. 
In such cases fine structures experience an over-contraction and leading to a distortion of their topological characteristics. 
In addition, LBC shows a topologically incorrect tree skeleton for trunk structures that have holes in the point cloud. 
Holes appear frequently due to occlusions and incomplete reconstructions. 
This incorrect contraction is caused by points at borders of holes. 
For the computation of the mean curvature flow the normal vector does not face perpendicular to the point surface but points in an almost vertical orientation. 
Due to the direction of the mean curvature flow, the contraction force points in the wrong direction, and elliptical artifacts appear in the skeleton. 
This effect is demonstrated in the middle column of Fig. \ref{fig:lbc_vs_slbc} and also in Fig. \ref{fig:skeleton_with_holes} in the appendix.

\begin{figure}[b]
    \centering
    \begin{subfigure}[t]{0.32\linewidth}
        \centering
       \includegraphics[width=\linewidth, trim=3 3 3 3,clip]{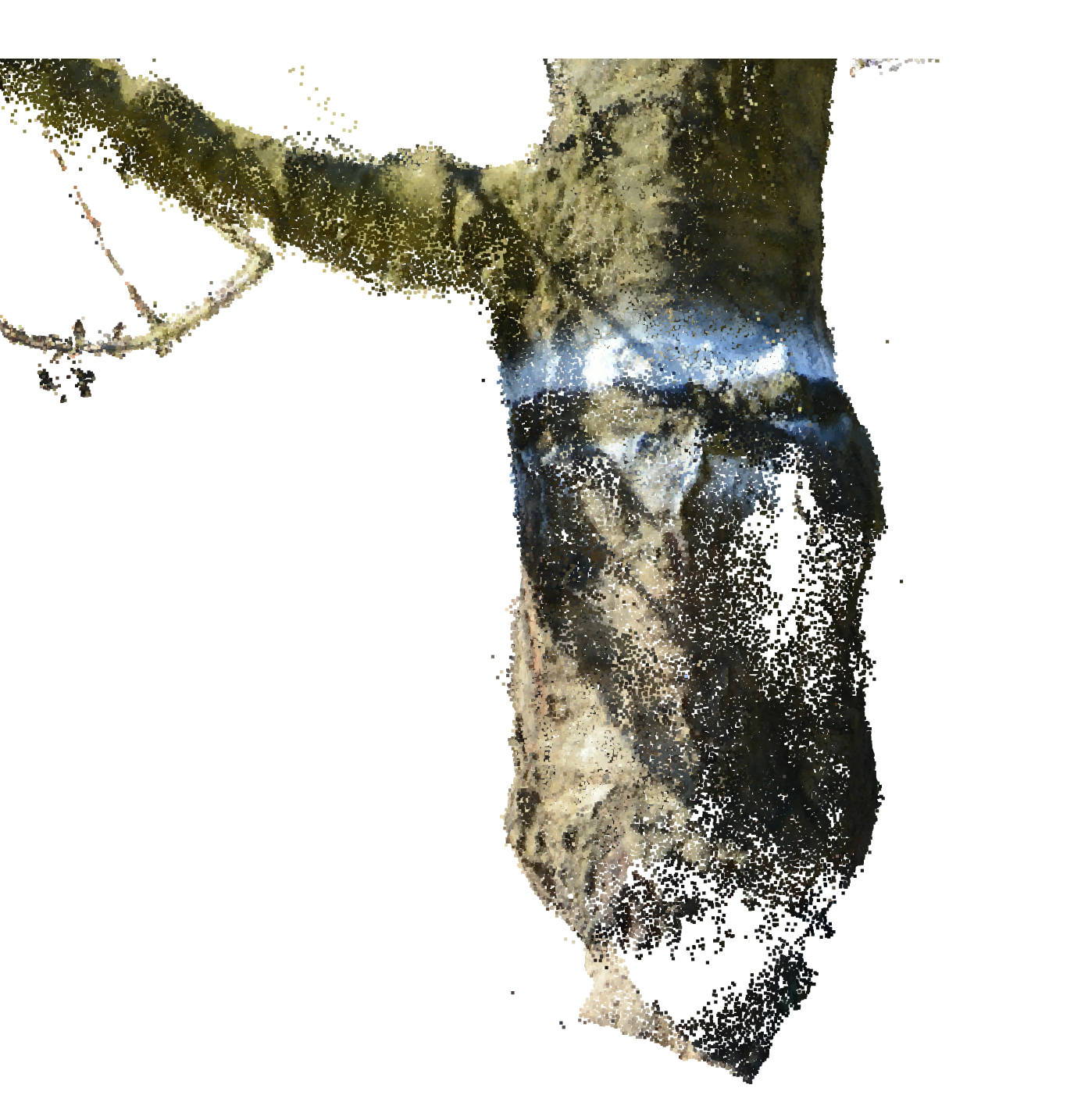}
       \caption{Original}
       \label{fig:lbc_vs_slbc:a}
    \end{subfigure}
    \begin{subfigure}[t]{0.32\linewidth}
        \centering
       \includegraphics[width=\linewidth, trim=3 3 3 3,clip]{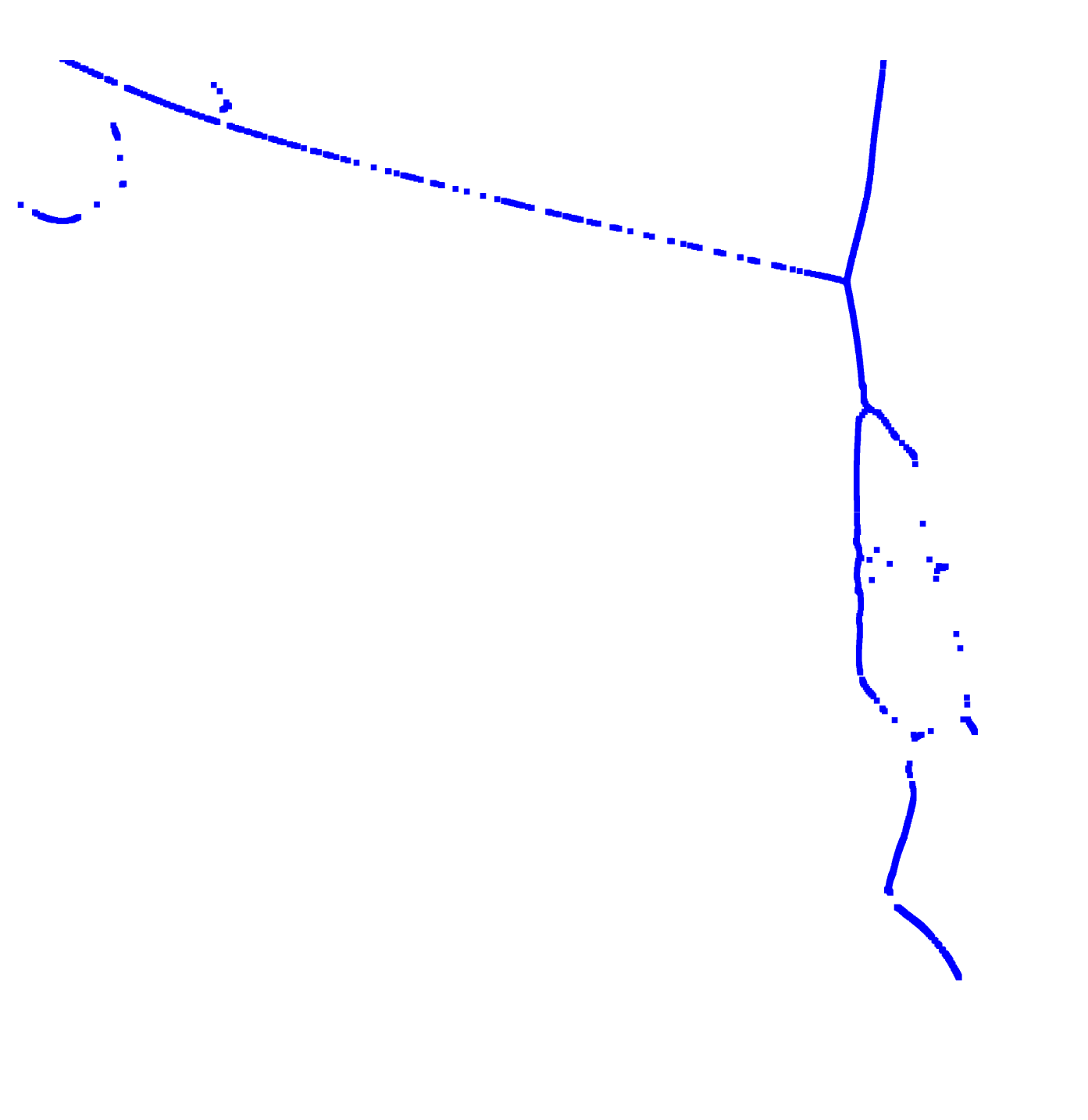}
       \caption{LBC}
       \label{fig:lbc_vs_slbc:b} 
    \end{subfigure}
    \begin{subfigure}[t]{0.32\linewidth}
        \centering
       \includegraphics[width=\linewidth, trim=3 3 3 3,clip]{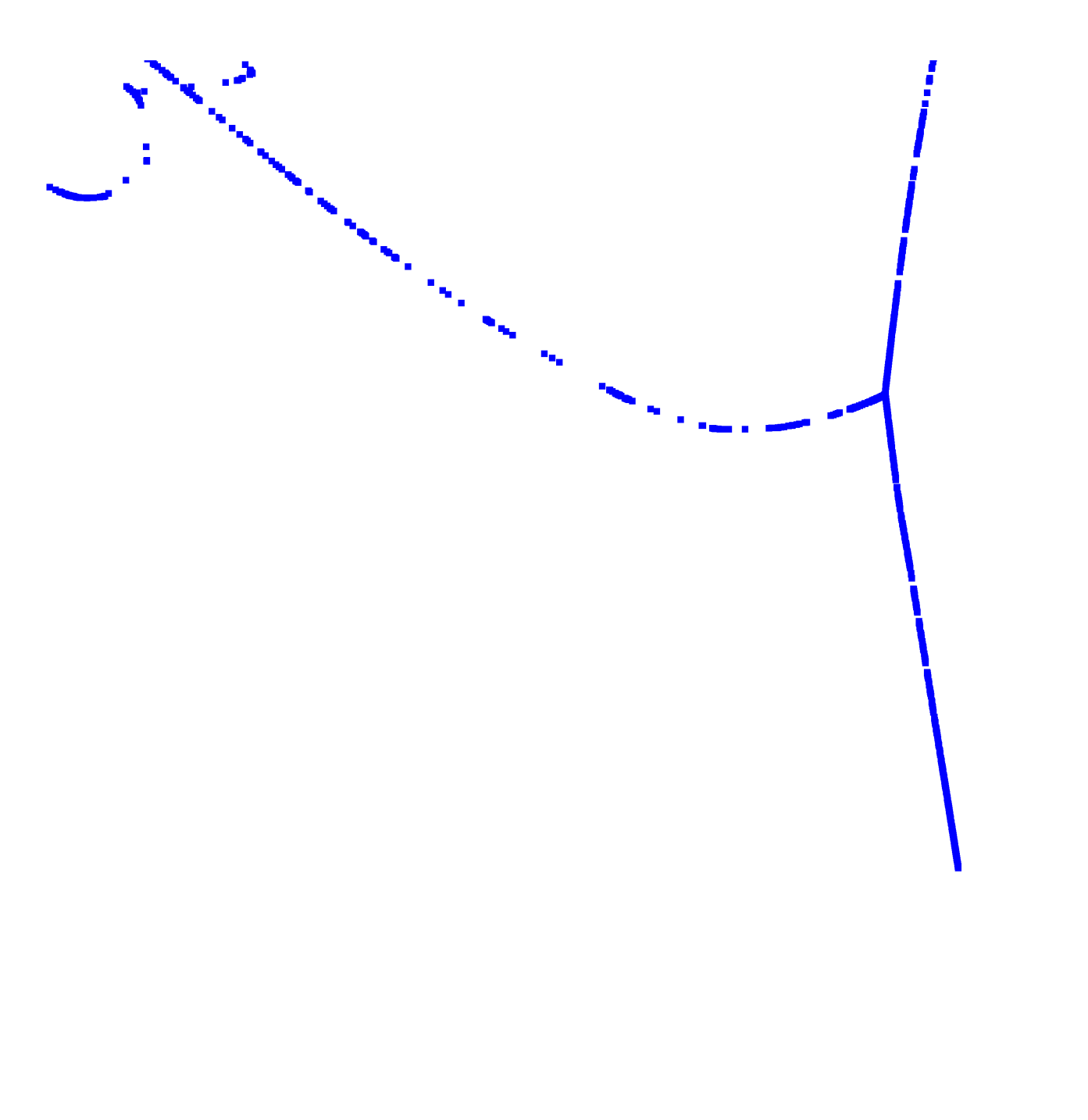}
       \caption{S-LBC (Ours)}
       \label{fig:lbc_vs_slbc:c}
    \end{subfigure}
    \caption{Comparison between Laplacian-based contraction (LBC) and the semantic Laplacian-based contraction (S-LBC) algorithm is shown on a real-world example with occluded areas in the original reconstruction. It is evident that S-LBC has a more plausible skeletonizing of the tree trunk than LBC. However, S-LBC exhibits over-smoothing at the joint of the branches.}
    \label{fig:lbc_vs_slbc}
\end{figure}
In order to address these topological artifacts, we introduce semantic Laplacian-based contraction (S-LBC). 
It integrates semantic information of the point cloud $\mathbf{C}_s$ into the LBC algorithm presented in Sec. \ref{ssec:pcsegmentation}. 
Even though the segmentation contains seven classes, we use only the \textit{trunk} and \textit{branch} classes. 
\begin{figure}[b]
    \centering
    \includegraphics[width=0.47\linewidth]{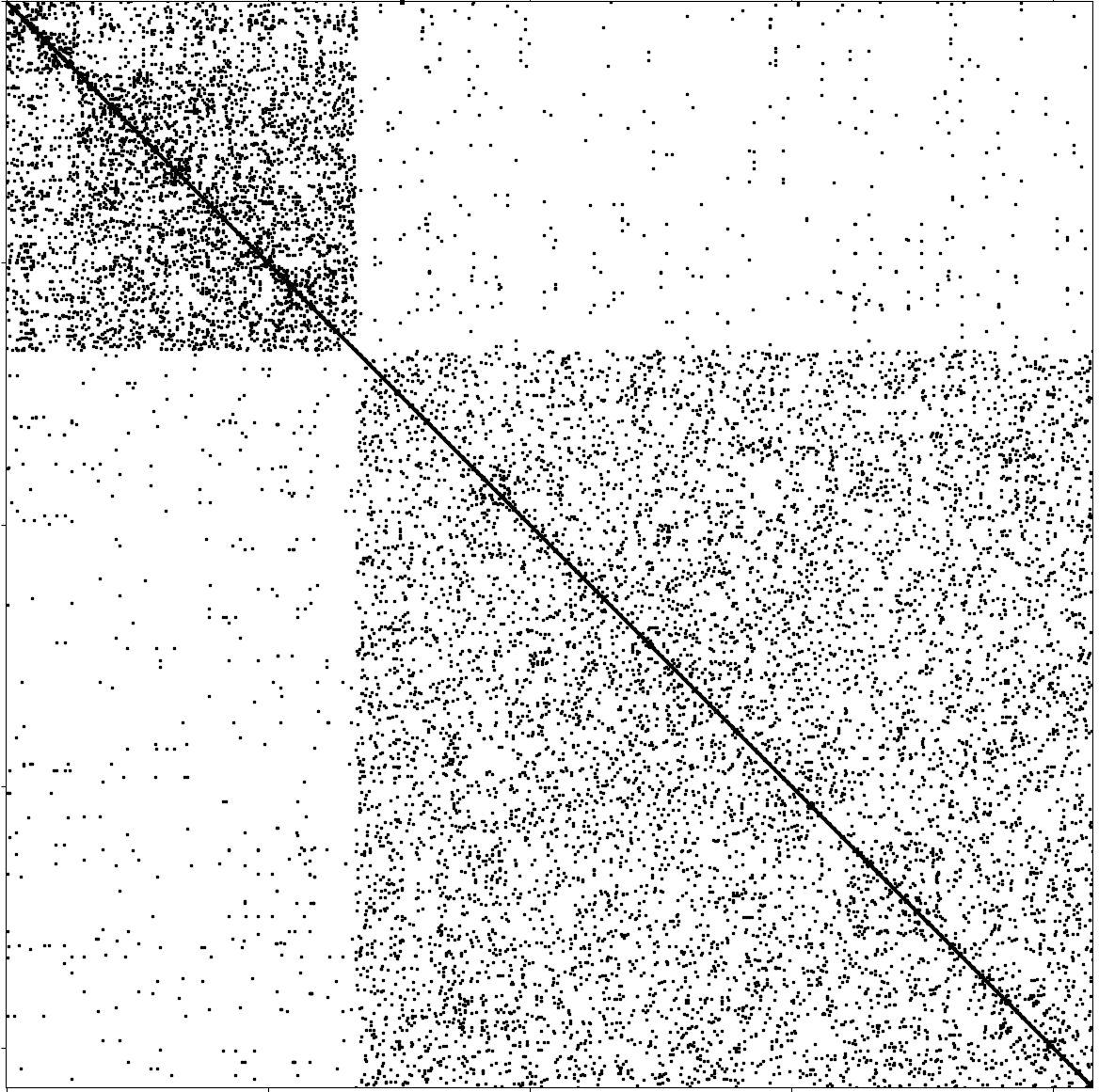}
    \caption{Illustration of the Laplacian matrix of the segmented point cloud $\mathbf{C}_s$ for both classes trunk and branch. The binary representation of the Laplacian matrix can be interpreted as the adjacency matrix of the point cloud. In this connectivity matrix (representing the points and their local and spatial vicinity) both submatrices for $\mathbf{L}_T$ (trunk) and $\mathbf{L}_B$ (branch) can clearly be separated. $\mathbf{L}_C$ describes the connection edges between both classes. For the semantic Laplacian-based contraction points having a connection in this area must be excluded, as it would distort the direction of the mean curvature flow.}
    \label{fig:laplacian_segmented_pc}
\end{figure}
They are represented as the trunk point cloud $\mathbf{C}_T$ and branch point cloud $\mathbf{C}_B$. In order to weight both classes differently, a semantic weighting matrix $\mathbf{S} \in \mathbb{R}^{N_C \times N_C}$ is defined.
\begin{figure*}[ht]
    \centering
    \begin{subfigure}[t]{0.27\linewidth}
       \includegraphics[width=\linewidth]{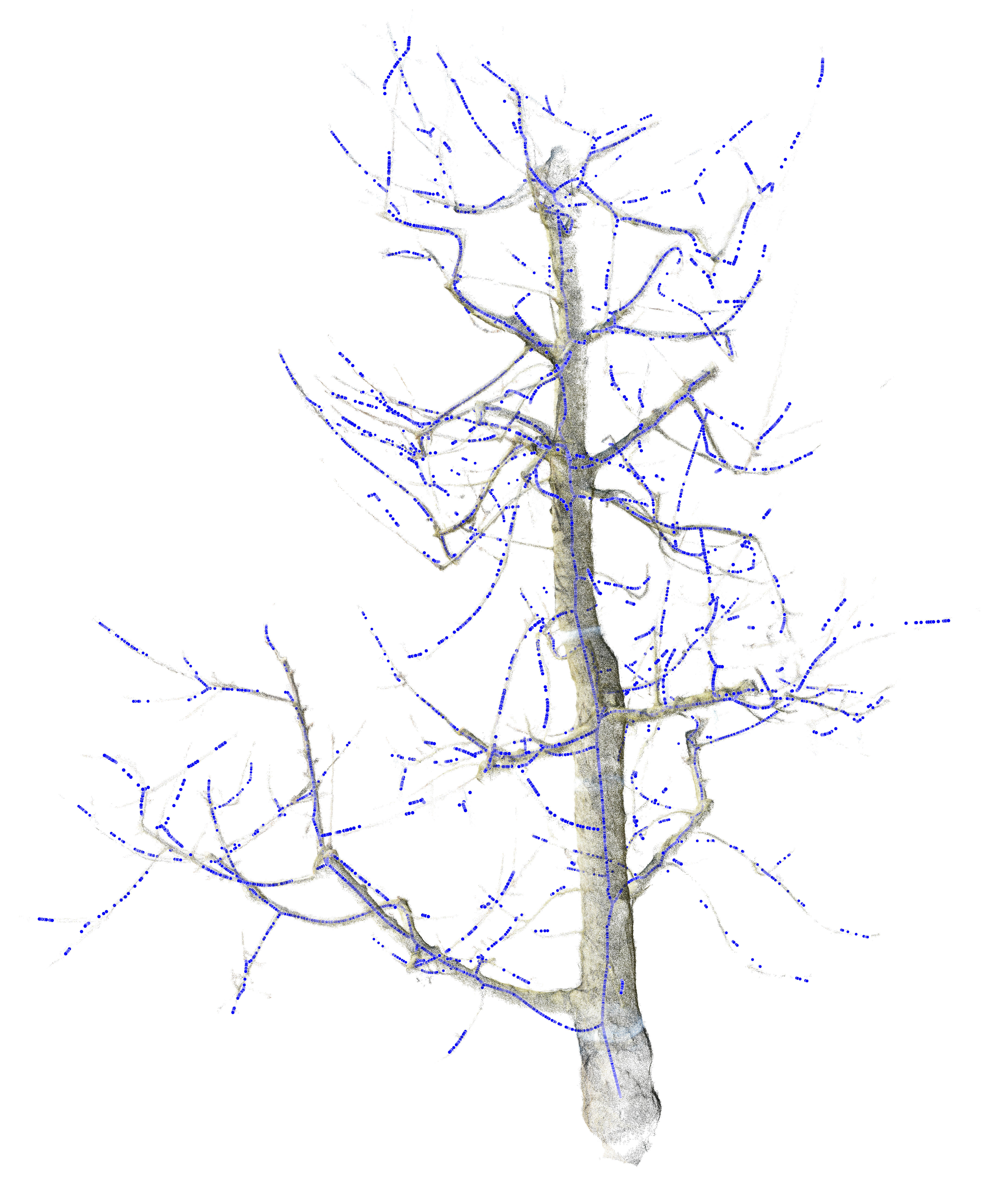}
       \caption{Original Point Cloud + Skeleton}
       \label{fig:results_skeleton:a} 
    \end{subfigure}
    \begin{subfigure}[t]{0.27\linewidth}
       \includegraphics[width=\linewidth]{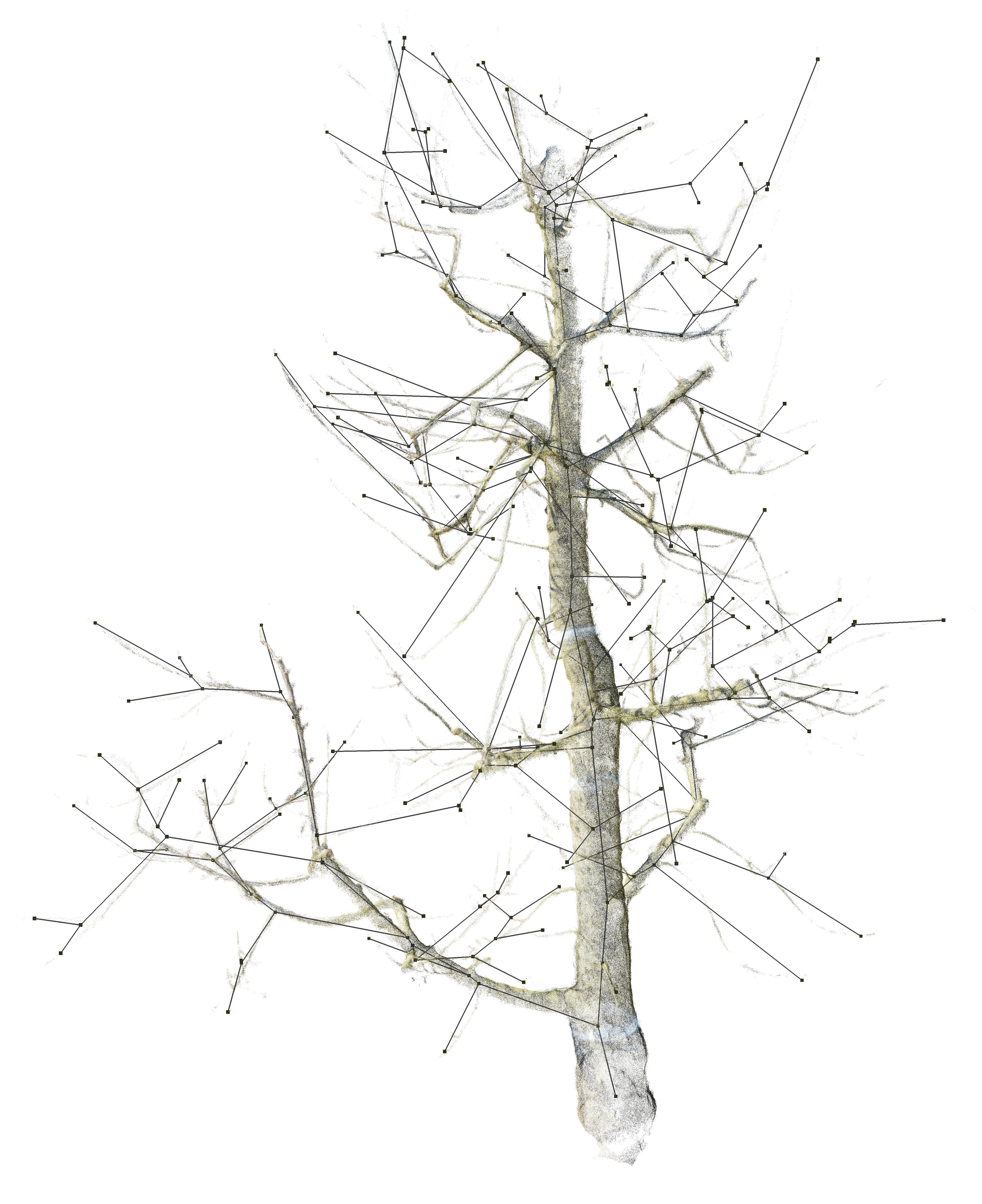}
       \caption{Tree Topology}
       \label{fig:results_skeleton:b}
    \end{subfigure}
        \begin{subfigure}[t]{0.22\linewidth}
       \includegraphics[width=\linewidth]{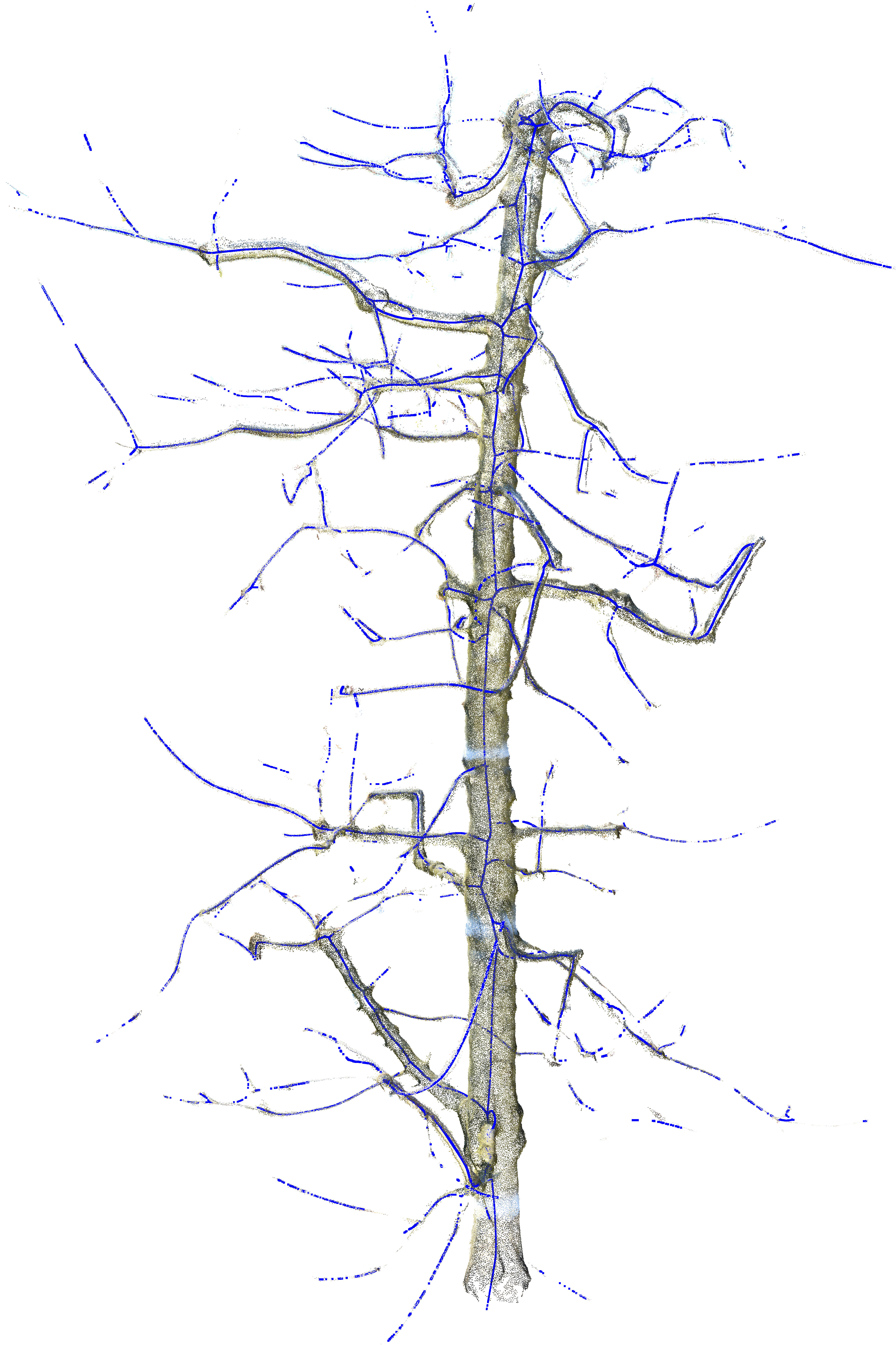}
       \caption{Original Point Cloud + Skeleton}
       \label{fig:results_skeleton:c} 
    \end{subfigure}
    \begin{subfigure}[t]{0.22\linewidth}
       \includegraphics[width=\linewidth]{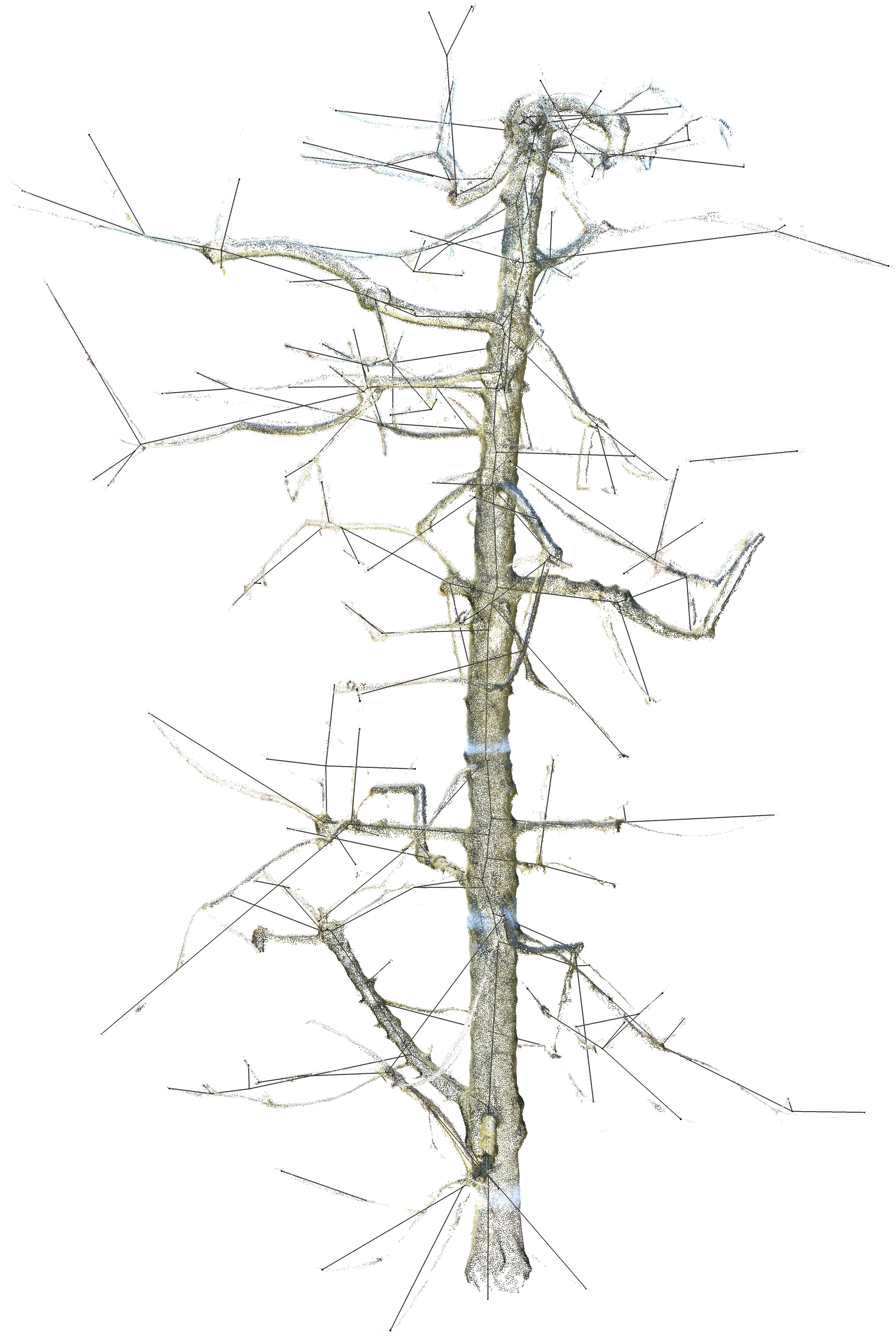}
       \caption{Tree Topology}
       \label{fig:results_skeleton:d}
    \end{subfigure}
\caption[]{Overlay of the skeleton and graph topology over real-world point cloud data. In both sub-figures Subfig. \ref{fig:results_skeleton:a} and \ref{fig:results_skeleton:c} the S-LBC skeleton point cloud from real tree data is displayed. Subfig. \ref{fig:results_skeleton:b} and \ref{fig:results_skeleton:d} represents the graph obtained from the minimum spanning tree.}
\label{fig:results_skeleton}
\end{figure*} 
The matrix $\mathbf{S}$ has a structural shape identical to that of the Laplacian matrix $\mathbf{L}$. 
Using the point-wise segmentation knowledge, we can efficiently make use of the Laplacian matrix $\mathbf{L}$ by rearranging it as follows:
\begin{equation}
    \mathbf{L} = 
    \begin{bmatrix}
\mathbf{L}_T & \mathbf{L}_C \\
\mathbf{L}^\top_C & \mathbf{L}_B
\end{bmatrix}.
\end{equation}
It can be seen that the Laplace matrix is divided into three distinct blocks. $\mathbf{L}_T \in \mathbb{R}^{N_T \times N_T}$, $\mathbf{L}_B \in \mathbb{R}^{N_B \times N_B}$, and $\mathbf{L}_C \in \mathbb{R}^{N_B \times N_T}$ contain the structural information about the trunk, the branch, and the connections between both classes, respectively. 
$N_T, N_B$ define the number of points in the trunk point cloud $\mathbf{C}_T$ and branch point cloud $\mathbf{C}_B$, respectively. 
A visualization of an example Laplacian matrix and its adjacency representation is depicted in Fig. \ref{fig:laplacian_segmented_pc}.
The semantic weighting matrix $\mathbf{S}$ has a similar structure. 
All connections of the trunk point cloud ($\mathbf{L}_T$) are weighted with $\lambda_T$ to exert an additional contraction force on the trunk. 
The weighting for the branch point cloud must remain constant and therefore we define $\lambda_B = 1$. 
Special attention has to be paid to the points which have a connection in the sub-Laplace matrix $\mathbf{L}_C$. 
These points are connecting pieces between the trunk and the branches and therefore have a connection with points from both classes. 
To get a topologically correct result all points sharing a neighboring point of the opposite class have to be weighted equally. 
Otherwise, the computation of the mean curvature flow is heavily distorted. 
To achieve this, all points which share a connection via the sub-Laplace matrix $\mathbf{L}_C$ are weighted uniformly with $\lambda_B$. Furthermore, the point cloud $\mathbf{C}_T$ and $\mathbf{C}_B$ are now transformed into the sets $\mathcal{C}_T$ and $\mathcal{C}_B$. This results in the following weighting rule for the elements $S_{ij}$ of $\mathbf{S}$:
\begin{equation}
S_{ij} =
    \begin{cases}
    \lambda_T & \text{if} ~ \mathbf{p}_i \in \mathcal{C}_T \wedge \mathcal{N}(\mathbf{p}_i) \nsubseteq\mathcal{C}_B,\\
    \lambda_B & \text{otherwise}
\end{cases}.
\end{equation}
This rule states that if a point $\mathbf{p}_i$ is a trunk point and its neighboring points $\mathcal{N}(\mathbf{p}_i)$ are not contained in the branch point cloud it is weighted with $\lambda_T$. To integrate the semantic weight matrix into the LBC Eq. \ref{eq:lbc},  $\mathbf{S}$ has to be multiplied element-wise with the right term $\mathbf{W}_L  \mathbf{L}$. Thus, the semantic Laplacian-based contraction(S-LBC) results in 
\begin{equation}
\begin{bmatrix}
\mathbf{S} \circ \mathbf{W}_L \mathbf{L}\\
\mathbf{W}_H
\end{bmatrix} \mathbf{C}^{'} =
\begin{bmatrix}
\mathbf{0}\\
\mathbf{W}_H \mathbf{C}
\end{bmatrix},
\end{equation}
where $\circ$ represents the Hadamard product. As a result the contracted point cloud skeleton $\mathbf{C}^{'}$ is obtained. It should be mentioned that S-LBC is not limited to only two classes but can be extended to any number of classes. Thus, different contraction weights can be assigned adaptively for different classes. 
\subsection{Graph Extraction}
\label{ssec:graph}
To decrease the number of skeletal points, we use farthest point sampling to down-sample the skeleton point clouds. 
Then, a minimum spanning tree is applied to the down-sampled skeletal points to obtain an undirected graph. 
As we are solely concerned with the branch junctions and endpoints, we simplify the graph by removing nodes with only two edges, leaving only nodes with three or more edges (representing junctions) and nodes with only one edge representing the start/end of the trunk or the tip of a branch. 
The output of this process is illustrated in Fig. \ref{fig:results_skeleton:b} and  Fig. \ref{fig:results_skeleton:d}.

Although the result appears satisfactory from a visual standpoint, close examination reveals that the graph may contain false connections if there is a large gap between the actual branch and the assigned branch. 
This issue is particularly prevalent at the treetop, where branches may touch and thus create incorrect graph connections. 
To address this problem, the point cloud density must be increased, which necessitates acquiring a greater number of images.

\section{Evaluation}

A straightforward evaluation of real data is not possible, since no reference data of a tree skeleton can be collected. In literature, the skeletons are evaluated based on visual appearance \cite{Laplacian, L1} or on synthetic 3D models of trees \cite{l1_mst, skeleton_pc_eval}. \cite{tree_dataset} published an open-source dataset for synthetic trees with ground truth skeletons but only contains sparse point clouds ($\sim 3000$ points). 

Therefore we divide the evaluation of our approach into two components: a visual and a systematic assessment. The visual evaluation will demonstrate the performance of S-LBC's on real-world data from the \textit{CherryPicker} pipeline, while the systematic evaluation will employ metrics to assess its performance on a synthetic tree dataset.\\

\noindent\textbf{Visual Evaluation}. During vegetation dormancy, we performed multiple 3D reconstructions of different cherry trees. For this evaluation, we applied the S-LBC on two different cherry tree data sets. The result for the skeletonization algorithm for the different trees is depicted in Fig. \ref{fig:results_skeleton}. The visual assessment shows that the skeletons are contracted with visually pleasing results. Missing data in the original point cloud lead to incomplete reconstruction and can only be eliminated by increasing the point cloud quality by using more images for the reconstruction. Artifacts like missing data led to gaps in the skeleton and might cause problems during the construction of the topology and can only be corrected by manually editing the created topology.\\

\noindent\textbf{Systematic Evaluation}. We generated a dataset of $50$ trees using \textit{Blender}, a 3D modeling tool, and its integrated tree generation tool \cite{blender}.
While creating the dataset, we made sure to generate trees resembling the structure of the cherry trees at our cherry orchard, i.e. with relatively thick trunks and thin branches. 
Therefore, templates were utilized for Douglas fir, larch, and pine. 
Each extracted tree model contains a mesh of the entire tree, a mesh of the tree trunk, and the ground truth skeleton represented as a line set. 
The meshes of the tree were uniformly sampled and we added Gaussian noise $\mathcal{N}(0,3 \sigma_d)$, where $\sigma_d$ represents the mean distance between the points of the point cloud. 
For evaluation, we computed the Chamfer distance (CD) \cite{chamfer_distance}. The CD finds for each point in the estimated point set $\mathcal X$, the nearest neighbor in the other ground truth set $\mathcal Y$, and averages the squared distances. The Chamfer distance is computed by
\begin{equation}
\begin{split}
	d_{CD}(\mathcal X,\mathcal Y) = &\frac{1}{|\mathcal X|} \sum_{\mathbf{x} \in \mathcal X} \min_{\mathbf{y} \in \mathcal Y}  ||\mathbf{x}-\mathbf{y}||^2_2 +\\
 &\frac{1}{|\mathcal Y|} \sum_{\mathbf{y} \in \mathcal Y} \min_{\mathbf{x} \in \mathcal X} ||\mathbf{x}-\mathbf{y}||^2_2
\end{split}.
\end{equation}
\begin{table}[b] 
\center
\caption{Evaluation of LBC and S-LBC. S-LBC outperforms LBC on noisy data with and without occlusions for both distance metrics. For comparison all parameters are set equally and only Laplacian weighting $\lambda_T=10$.}
\begin{tabular}{c| c} \toprule
Skeleton Algorithm   & Chamfer  \\ \midrule 
LBC with noise  & $37.555$ \\ 
S-LBC  with noise & $\mathbf{25.349}$\\ \midrule	
LBC  with noise \& occlusion  & $51.072$ \\ 
S-LBC with noise \& occlusion & $\mathbf{26.353}$ \\  \bottomrule	
\end{tabular}
\label{tab:skeleton_result}
\end{table}
In order to achieve a uniform density and realistic run times, the point clouds were down-sampled on a voxel grid with a size of $1.5$\,cm. For the first evaluation, the noisy data set was processed by both algorithms. The visual results are shown in the appendix in Fig. \ref{fig:skeleton_without_holes}.
and an evaluation of both metrics is shown in Tab. \ref{tab:skeleton_result}. 
S-LBC performs better on the tree dataset regarding the CD. 
This can be explained by the fact that in the region of the tree trunk, the skeleton is smoothed to a greater extent and thus lies closer to the ground truth skeleton. 
In the second evaluation, we assessed the performance of both algorithms on noisy and occluded data. 
To simulate this scenario, we randomly selected multiple points on the tree trunk and inserted holes with a radius of $8$\,cm around them. 
The resulting point clouds with holes are shown in the appendix in Fig. \ref{fig:skeleton_with_holes}. 
The skeletons generated from LBC, shown in the middle column of Fig. \ref{fig:skeleton_with_holes} in the appendix
, artifacts appear on the skeleton at the position of the occlusion. 
It can be seen that with S-LBC, the artifacts disappear or will be attenuated due to semantic weighting. 
The result of Tab. \ref{tab:skeleton_result} reflects this improvement and the Chamfer distance stays almost identical for S-LBC compared to the noise-only evaluation. 
On the other side, LBC suffers from occlusions which is reflected in the CD.

\section{Conclusion \& Future Work}

This work presents \textit{CherryPicker}, an automated system pipeline for obtaining the topology of cherry trees. Within this pipeline, we introduced an algorithm to automatically determine the scale factor of a monocular 3D reconstruction by placing an ArUco marker with a known size into the scene. To extract the tree's topology, we proposed a semantic Laplacian-based contraction algorithm that incorporates semantic knowledge of segmented tree point clouds. In our case, the weighting differs between branches and tree trunk. However, the semantic weighting can be extended to multiple classes with individual weighting. In a visual evaluation, we showed that the algorithm produces high-quality skeletons. Additionally, the systematic evaluation showed improved results regarding occlusions and structural disparity variations.

One limitation of utilizing real data is the absence of benchmarks to assess the efficiency and quality of the algorithms. 
Therefore, we plan to create an artificial, photo-realistic tree dataset containing comprehensive 2D and 3D semantic information, depth maps, and skeleton ground truths. 
This dataset can than be used to implement a learning-based approach for extracting the point cloud skeletons, inspired by the method introduced in \cite{Point2Skeleton}. 
An additional constraint is that the current methodology only permits the skeletonization of tree structures without leaves. 
However, in future work, it may be feasible to conduct the scan during vegetation dormancy and fit the skeleton into a 3D reconstruction with leaves. 
This approach would enable the intra-seasonal use of the skeleton data, facilitate the assignment and thus allow the integration of leaves and fruits into the existing skeleton.

\section{Acknowledgement}
We thank \textbf{Max Weiherer}, \textbf{Vanessa Wirth} and \textbf{Jan-Ole Henningson} for their valuable feedback. This project is funded by the 5G innovation program of the German Federal Ministry for Digital and Transport under the funding code 165GU103B.

{\small
\bibliographystyle{ieee_fullname}
\bibliography{library}
}

\section*{Appendix}

\begin{figure*}
    \centering
    \def\svgwidth{0.95\linewidth}
    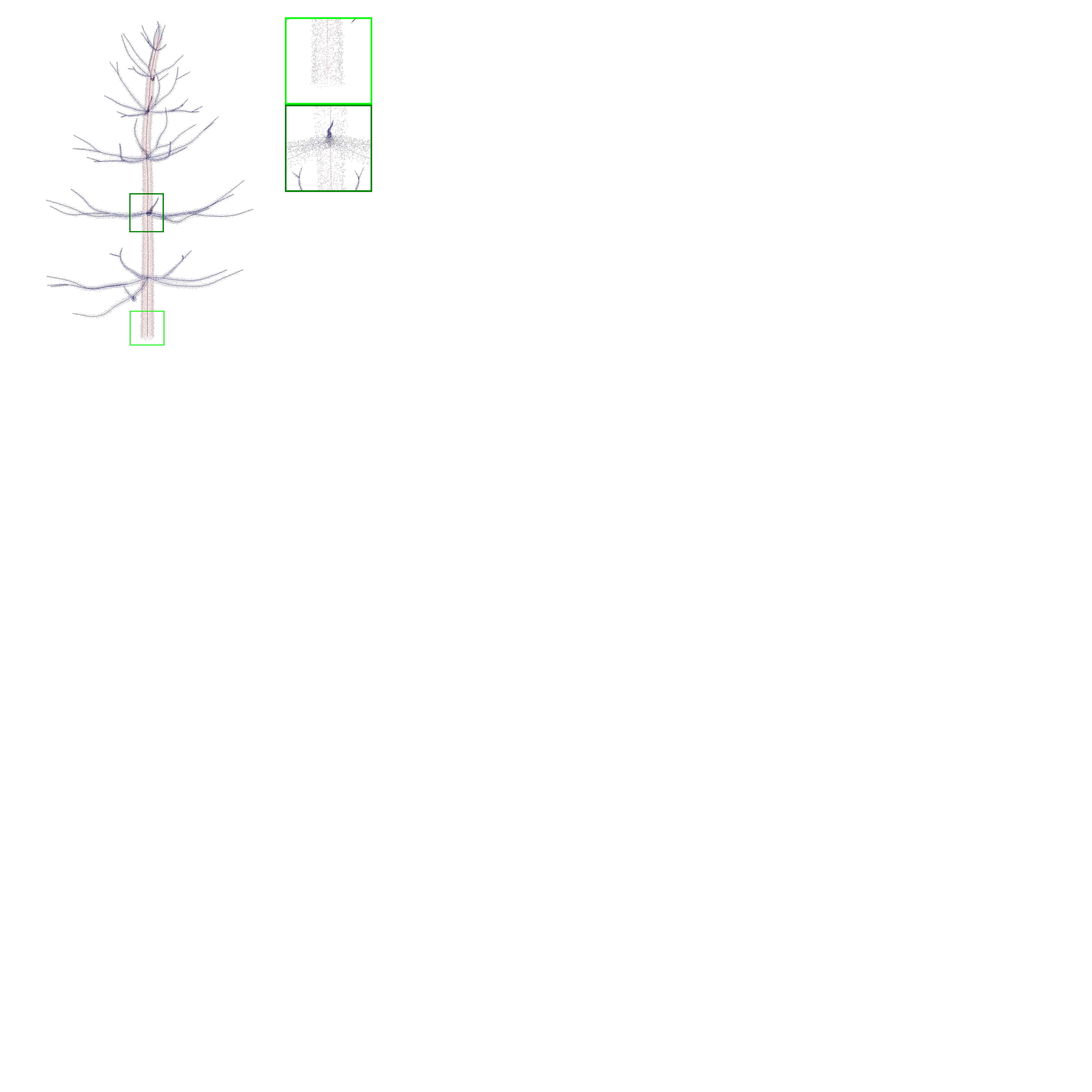
    \caption{For the visualization of the advantages of semantic Laplacian-based contraction (S-LBC) over standard Laplacian-based contraction (LBC) we show different cases. With the larch, we chose balanced contraction $\mathbf{W}_L$ and attraction weights $\mathbf{W}_H$ for LBC. It is a compromise for a good reconstruction of thin branches and thick trunks. Nevertheless, LBC struggles to properly reconstruct the bottom of the trunk. S-LBC achieves the same results for the branches but tackles the artifact of the trunk. 
    In the pine example, we increased the contraction weights and decreased the positional weighting to tackle the trunk artifact within LBC. It can be seen that LBC eliminates the trunk artifacts but introduces over-smoothing at the branches. For S-LBC the skeleton benefits from different weighting and has no artifacts. 
    At the last example executed at a fir, LBC weighting for $\mathbf{W}_H$ increased and decreased for $\mathbf{W}_L$. At the branches, the topological results are similar to the ground truth but the trunk suffers from an incorrect trunk skeleton. On the other side, S-LBC keeps the details of the branch and additionally contracts the trunk properly.}
    \label{fig:skeleton_without_holes}
\end{figure*}

\begin{figure*}
    \centering
    \def\svgwidth{0.95\linewidth}
    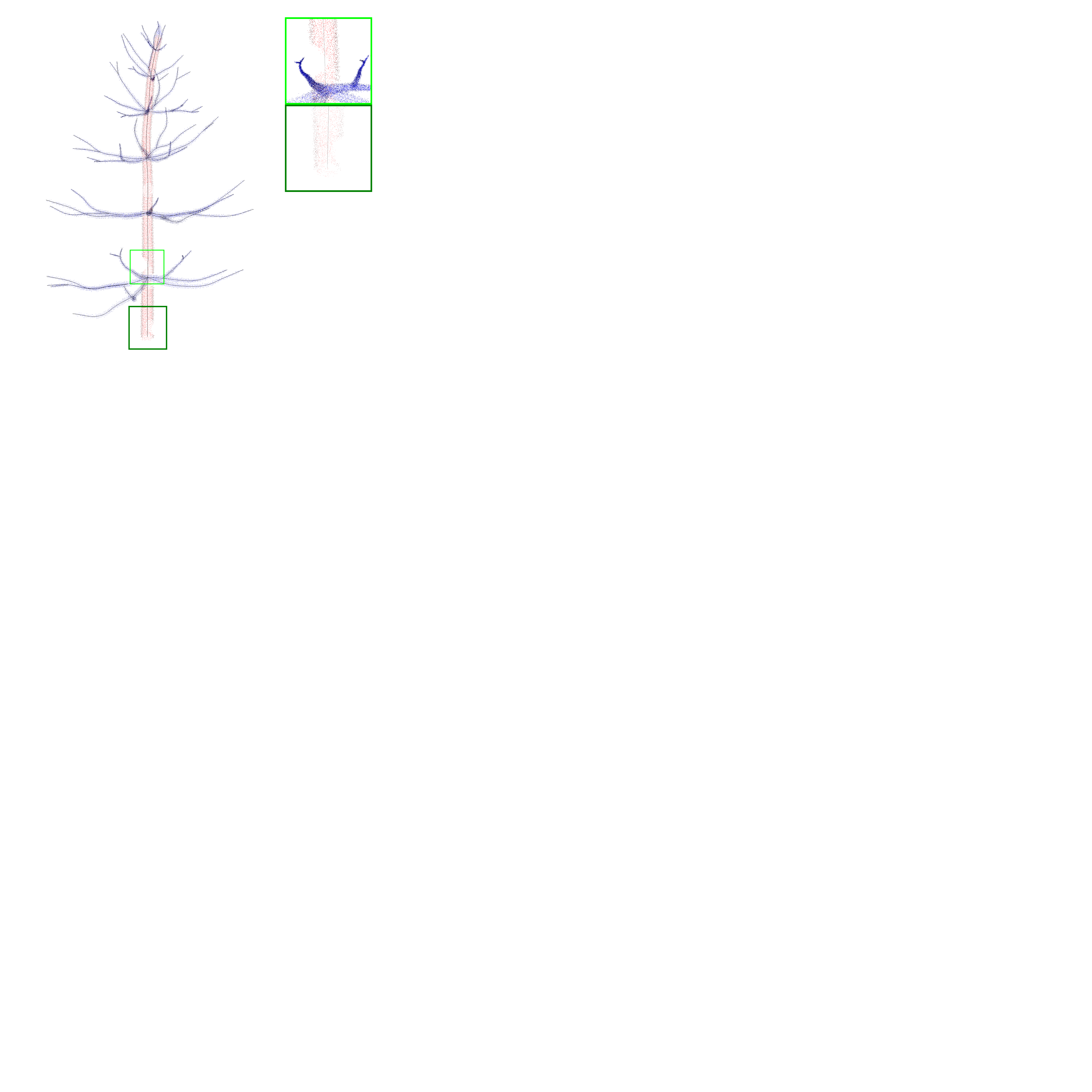
    \caption{Visualization of the skeleton algorithms LBC and S-LBC on noisy and occluded tree point cloud data. In the region of holes, the Laplacian-based contraction (LBC) shows circular artifacts. For the different tree types, we choose similar contraction $\mathbf{W}_L$ and attraction weights $\mathbf{W}_H$. At the top row with the larch example, it can be seen LBC shows elliptical errors at the bottom tree trunk. The green area shows a curved skeleton for LBC and for S-LBC it is a straight line due to the additional trunk weighting.
    In the pine example, it can be seen LBC shows erroneous skeletons in the trunk and additionally missing links appear. S-LBC is able to counteract these artifacts.
    Lastly, in the Fir example, the holes produce different artifacts for LBC. The S-LBC is able to overcome these artifacts by enforcing stronger contractions due to semantic weighting.}
    \label{fig:skeleton_with_holes}
\end{figure*}

See next pages for Fig. \ref{fig:skeleton_without_holes} and Fig. \ref{fig:skeleton_with_holes}.

\end{document}